# Improving parameter learning of Bayesian nets from incomplete data

G. Corani and C. P. De Campos

IDSIA

Galleria 2, CH-6928 Manno (Lugano)

Switzerland

{giorgio,cassio}@idsia.ch

**Abstract**

This paper addresses the estimation of parameters of a Bayesian network from incomplete data. The task is usually tackled by running the Expectation-Maximization (EM) algorithm several times in order to obtain a high log-likelihood estimate. We argue that choosing the maximum log-likelihood estimate (as well as the maximum penalized log-likelihood and the maximum a posteriori estimate) has severe drawbacks, being affected both by overfitting and model uncertainty. Two ideas are discussed to overcome these issues: a maximum entropy approach and a Bayesian model averaging approach. Both ideas can be easily applied on top of EM, while the entropy idea can be also implemented in a more sophisticated way, through a dedicated non-linear solver. A vast set of experiments shows that these ideas produce significantly better estimates and inferences than the traditional and widely used maximum (penalized) log-likelihood and maximum a posteriori estimates. In particular, if EM is adopted as optimization engine, the model averaging approach is the best performing one; its performance is matched by the entropy approach when implemented using the non-linear solver. The results suggest that the applicability of these ideas is immediate (they are easy to implement and to integrate in currently available inference engines) and that they constitute a better way to learn Bayesian network parameters.

## 1 Introduction

This paper focuses on learning the parameters of a Bayesian network (BN) with *known structure* from incomplete samples, under the assumption of MAR (missing-at-random) missing data. In this setting, the missing data make the log-likelihood (LL) function non-concave and multimodal. The most common approach to maximize LL in the presence of missing data is the Expectation-Maximization (EM) algorithm [4], which generally converges to a local maximum of the LL function. The EM can be easily modified to maximize, rather than LL, the posterior probability of the data (MAP), as well as other *penalized maximum likelihood* ideas [11, Sec 1.6]. Generally, maximizing MAP rather than LL yields smoother estimates, less prone to overfitting [8]. In the following, we refer to the function to be maximized as the *score*. Although we focus on BN learning, the ideas are general and shall apply also to other probabilistic graphical models that share similar characteristics in terms of paramater learning.

In order to reduce the chance of obtaining an estimate with low score, a multi-start approach is adopted: EM is started from many different initialization points, eventually selecting the estimate corresponding to the run that achieves the highest score. We argue however that this strategy has some drawbacks. Firstly, the estimate that maximizes the score can well do so because of overfitting. Even if the MAP estimation is adopted, the fixed structure and the amount of data may lead to overfitting, because they might not fully represent the distribution that generated the data. Secondly, the estimates produced by the different EM runs are typically very different from each other, and yet achieve very close scores [7, Chap. 19]. Choosing the single estimate with highest score implies in *model uncertainty*, because the estimates with slightly worse score are completely ignored. Overall, the score alone does not seem to be powerful enough to identify the best estimate. Note that the challenge presented here does not involve model complexity: all the competing estimates have the same underlying structure and thus approaches such as the Bayesian Information Criterion (BIC) do not apply.

We propose and compare two approaches to replace the criterion of selecting the highest score estimate, both based on well-known ideas already applied in other contexts. The first is based on the principle of maximum entropy and the second on model averaging. The maximum entropy criterion can be stated as: "*when we make inferences on incomplete information, we should draw them from that probability distribution that has the maximum entropy permitted by the information which we do have*" [6]. We interpret this principle by first discarding estimates with low score, which we assume to be poor, and then by selecting the most entropic estimate among the remaining ones. The entropy-based criterion is expected to yield parameter estimates that are more robust to overfitting than those from the criterion of maximum score. In its simplest version, we apply the entropy principle on top of a multi-start EM; in a more sophisticated fashion, we implement it using a non-linear solver. In [12], a similar idea has been explored to fit continuous distributions from only partial knowledge about moments or other features that are extracted from data. The model averaging idea is instead inspired by Bayesian Model Averaging (BMA, see [5]), and is designed to be used on top of the multi-start EM. BMA is a technique specifically designed to deal with model uncertainty, which prescribes to average the predictions produced by a set of competing models, assigning to each model a weight proportional to its posterior probability. In the literature, BMA has been often used to manage ensembles of BN classifiers; yet, most attempts were not successful, as reviewed by [2]. The problem is that a single model becomes usually much more probable than any competitor, and thus there is little difference between using BMA or the single most probable model (that is, the maximum score one). Our approach is to apply BMA locally to each conditional probability distribution that defines the BN; this allows us to instantiate a single BN model, whose parameters can be readily inspected, instead of dealing with a collection of models.

Sections 3 and 4 present a vast amount of experiments with different BNs and missingness processes. These experiments, performed on a grid of computers, would have taken more than nine months if run on a single desktop computer. They consistently show that both the entropy and the BMA-based approaches provide better estimates than the maximum score estimation, and that these better parameters also result in better inferences with the resulting BNs.

## 2 Methods

We adopt Bayesian networks as framework for our study, even if the discussion of this paper may be relevant to the parameter learning of other probabilistic graphical models. Therefore, we assume that the reader is familiar with basic concepts of Bayesian networks [7]. A Bayesian network (BN) is a triple $(\mathcal{G}, \mathcal{X}, \mathcal{P})$, where $\mathcal{G}$ is a directed acyclic graph with nodes associated to random variables $\mathcal{X} = \{X_1, \ldots, X_n\}$ over discrete domains $\{\Omega_{X_1}, \ldots, \Omega_{X_n}\}$ and $\mathcal{P}$ is a collection of probability values $p(x_j|\pi_j)$ with $\sum_{x_j \in \Omega_{X_j}} p(x_j|\pi_j) = 1$, where $x_j \in \Omega_{X_j}$ is a category of $X_j$ and $\pi_j \in \times_{X \in \Pi_j} \Omega_X$ an instantiation for the parents $\Pi_j$ of $X_j$ in $\mathcal{G}$. Furthermore, every variable is conditionally independent of its non-descendants given its parents. Given its independence assumptions, the joint probability distribution represented by a BN is obtained by $p(\mathbf{x}) = \prod_j p(x_j|\pi_j)$, where $\mathbf{x} \in \Omega_{\mathcal{X}}$ and all $x_j, \pi_j$ (for every $j$) agree with $\mathbf{x}$. Uppercase letters are used for random variables and lowercase letters for their corresponding categories. The graph $\mathcal{G}$ and the variables $\mathcal{X}$ (and their domains) are assumed to be known; $\theta_{\mathbf{v}|\mathbf{w}}$ is used to denote the probability $p(\mathbf{v}|\mathbf{w})$ (with $\mathbf{V}, \mathbf{W} \subseteq \mathcal{X}$).

Given the data $\mathbf{y} = (\mathbf{y}^1, \ldots, \mathbf{y}^N)$ with $N$ instances such that $\mathbf{y}^i \in \Omega_{\mathbf{Y}^i}$ and $\mathbf{Y}^i \subseteq \mathcal{X}$ is the set of observed variables of instance $i$, we denote by $N_{\mathbf{w}}$ the number of instances of $\mathbf{y}$ that agree with the configuration $\mathbf{w}$. The goal is to learn $\mathcal{P}$, which is usually done by maximizing the penalized (log-)likelihood (LL) of $\mathbf{y}$:

$$\hat{\boldsymbol{\theta}} = \underset{\boldsymbol{\theta}}{\operatorname{argmax}}\, S_{\boldsymbol{\theta}}(\mathbf{y}) = \underset{\boldsymbol{\theta}}{\operatorname{argmax}} \left( \sum_{i=1}^{N} \log \theta_{\mathbf{y}^i} + \alpha(\boldsymbol{\theta}) \right),$$

where $\alpha$ is the penalty term. The argument $\mathbf{y}$ of $S_{\boldsymbol{\theta}}$ is omitted from now on ($S$ is the acronym for *score*). We use the penalized LL because one might simply set the penalty to zero to obtain the standard likelihood, or to $\log p(\boldsymbol{\theta})$ to get the MAP estimation. For ease of expose, we assume:

$$\alpha(\boldsymbol{\theta}) = \log \prod_{j=1}^{n} \prod_{x_j} \prod_{\pi_j} \theta_{x_j|\pi_j}^{\alpha_{x_j,\pi_j}},$$

with $\alpha_{x_j,\pi_j} = \frac{1}{|\Omega_{X_j}| \cdot |\Omega_{\Pi_j}|}$, which is the MAP version with equivalent sample size set to one. For a complete data set (that is, $\mathbf{Y}^i = \mathcal{X}$ for all $i$), we have a concave sum of logarithms on $\boldsymbol{\theta}$:

$$S_{\boldsymbol{\theta}} = \sum_{j=1}^{n} \sum_{x_j} \sum_{\pi_j} N'_{x_j,\pi_j} \log \theta_{x_j|\pi_j},$$

where $N'_{x_j,\pi_j} = N_{x_j,\pi_j} + \alpha_{x_j,\pi_j}$, and the estimate $\hat{\theta}_{x_j|\pi_j} = N'_{x_j,\pi_j}/(\sum_{x_j} N'_{x_j\pi_j})$ achieves maximum score. In the case of incomplete data, we have

$$S_{\boldsymbol{\theta}} = \sum_{i=1}^{N} \log \sum_{\mathbf{z}^i} \prod_{j=1}^{n} \theta_{x_j^i|\pi_j^i} + \alpha(\boldsymbol{\theta}), \qquad (1)$$

where $\mathbf{x} = (\mathbf{y}^i, \mathbf{z}^i) = (x_1^i, \ldots, x_n^i)$ represents an instantiation of all the variables. No closed-form solution is known, and one has to directly optimize $\max_{\boldsymbol{\theta}} S_{\boldsymbol{\theta}}$, subject to $\forall_j \forall_{\pi_j} : \sum_{x_j} \theta_{x_j|\pi_j} = 1$, $\forall_j \forall_{x_j} \forall_{\pi_j} : \theta_{x_j|\pi_j} \geq 0$.

The most common approach to optimize this function is to use the EM method, which completes the data with the expected counts for each missing variable given the observed variables, that is, variables $Z_j^i$ are completed by "weights" $\hat{\theta}_{Z_j|\mathbf{y}^i}^k$ for each $i, j$ of a missing value, where $\hat{\boldsymbol{\theta}}^k$ represents the current estimate at iteration $k$. This idea is equivalent to weighting the chance of having $Z_j^i = z_j$ by the (current) distribution of $Z_j$ given $\mathbf{y}^i$ (this is known as the *E-step*, and requires inferences over the network instantiated with $\mathcal{P} = \hat{\boldsymbol{\theta}}^k$). Using these weights together with the actual counts from the data, the sufficient statistics values $N_{x_j,\pi_j}^k$ are computed for every $x_j, \pi_j$, and the next (updated) estimate $\hat{\boldsymbol{\theta}}^{k+1}$ is obtained as if the data were complete: $\hat{\theta}_{x_j|\pi_j}^{k+1} = N'^k_{x_j,\pi_j}/(\sum_{x_j} N'^k_{x_j\pi_j})$, where $N'^k_{x_j,\pi_j} = N^k_{x_j,\pi_j} + \alpha_{x_j,\pi_j}$ as before (this is the *M-step*). Because in the first step there is no current estimate $\hat{\boldsymbol{\theta}}^0$, an initial guess has to be used. Using the score to test convergence, this procedure achieves a saddle point, which is usually a local optimum of the problem, and may vary according to the initial guess $\hat{\boldsymbol{\theta}}^0$. In view of obtaining an estimate with high score, it is common to execute multiple runs of EM with distinct initial guesses and then to take the estimate with maximum score among them. However, it is often the case that many distinct estimates have very similar score, and simply selecting the one with the highest one is clearly an over-simplified decision, because equal (or almost equal) scores cannot be used as a criterion to find the best estimate.

## 2.1 Entropy

High score is not the only target in order to obtain a good estimate. There are many estimates that lie within a tiny distance from the global maximum and can be as good as or better than the global one. A simple alternative approach is to pick the parameter estimate which has *maximum entropy*, among those which have a high score [6]. Therefore, a possible criterion is

$$\hat{\boldsymbol{\theta}} = \operatorname*{argmax}_{\boldsymbol{\theta}} \sum_{j=1}^{n} \sum_{\pi_j} \sum_{x_j} \theta_{x_j|\pi_j} \log \theta_{x_j|\pi_j}, \qquad (2)$$

subject to $S_{\boldsymbol{\theta}} \geq c \cdot s^*$, for a given $0 \leq c \leq 1$, where $s^*$ is the maximum score. The optimization of Eq. (2) computes the maximum entropy distribution with the guarantee that its score is high (within a small difference from the maximum value). Eq. (2) is referred to as *local entropy* in [10].

The maximum entropy is used because it leads to the most conservative (least informative) distribution over the set of all estimates that achieve score as good as $c \cdot s^*$. The constraint that bounds the score is not as simple as it reads: in fact it is necessary to use all the equations that define the score function to force it to be greater than a certain value, if one wants to use a non-linear optimization suite. On the other hand, a simple implementation of entropy can be done by using the many runs of the EM method. The idea is to select, among the estimates returned by the different EM runs that achieve a high enough score (compared to the maximum obtained one), the estimate with maximum entropy. This differs from the usual maximum entropy inference in the way that it first checks for high score estimates and then maximizes entropy among them. Given the usual great number of parameters to estimate in a Bayesian network, restricting ourselves only to those estimates that have exactly equal score is undesired: usually only the top scoring estimate will be left.

A related maximum entropy approach is shown in [12], with two main differences: (i) they focus on a continuous scenario with a compact parametrization, e.g. using mean/variance or other similar features of the data, which implies in more estimates that equally fit the data (our setting has categorical data and there are tens to hundreds of parameters to estimate); (ii) they force their estimator to have likelihood precisely equal to the value of the maximum score (which in our case would be something similar to using $c = 1$). We note that they have also extended their idea to a so-called *regularized* version, which includes a penalty in the entropy function. This is shown to become similar to the MAP estimation. We work differently by allowing some variation in the score without the use of an extra penalization for that purpose, as we consider all estimates with high score as equally good (they are later discriminated by their entropy). In a BN with more than a couple of variables, the number of parameters to estimate becomes quickly large and there is only a very small (or no) region of the parameter space with estimates that achieve the very same global maximum value. However, a feasibility region defined by a small percentage away from the maximum score is enough to produce a whole region of estimates, indicating that the region of high score estimates is almost (but not exactly) flat. This is expected in a high-dimensional parameter space.

## 2.2 Bayesian Model Averaging

A BMA-based approach can also be used to overcome the model uncertainty and overfitting problems. The BMA is applied on the alternative estimates returned by the various EM runs in order to obtain a final single estimate. The rationale is as follows: if we consider as query the variable $X_j$ given its parents, the posterior probability distribution $p(X_j|\pi_j)$ returned by an inference corresponds to the distribution which is specified in the conditional probability table of the BN for $X_j$. To answer this query using BMA, we use the estimates identified by each run of EM. We average the returned inferences from models identified by each EM run, using weights proportional to the score achieved by them.[1] We repeat this query for each $j$ and each combination of $\pi_j$; eventually we instantiate a single BN, by setting $\hat{\theta}_{x_j|\pi_j} = p(x_j|\pi_j)$, where $p(x_j|\pi_j)$ is the BMA-averaged inference. In practice, this can be done by simply averaging the coefficients of the conditional probability tables. Thus, BMA is *locally* applied to estimate each conditional probability distribution; we denote as $\hat{\boldsymbol{\theta}}$ the estimate obtained in this way. Note that the inferences returned by $\hat{\boldsymbol{\theta}}$ match those produced by the standard BMA (which always computes the answer by querying over all the estimated models and then averaging the returned values) only for queries on $X_j$ given its parents, and not on more general queries. However, it is generally not possible to obtain a summary estimate which exactly matches the inferences produced by the standard usage of BMA. In fact, the standard BMA can be seen as an ensemble of BNs; to get a single estimate, one has to average the joint distributions of these BNs. This would produce a representation of the joint probability distribution that is not guaranteed to factorize as the original BN structure. Moreover, it would be very demanding from the computational viewpoint. An exception exists for naive structures [3], while our BMA-based approach can be used to estimate the parameters of *any* BN.

## 3 Experiments with EM as Underlying Engine

In order to compare entropy and BMA approaches, we perform experiments using different network structures (Asia, Alarm and randomly generated networks), sample sizes ($n$=100, $n$=200) and percentages of missing data $mp$ ($mp$=30%, $mp$=60%). We also include the maximum score approach in the experiments to serve as baseline, which we refer to as MAP (this is similar to a penalized LL, as defined in Sec. 2). A triple $\langle$network structure$, n, mp\rangle$ identifies a *setting*. For each setting, we perform 300 experiments, where each experiment is organized as follows: a) random draw of the parameters of the *reference network*; b) sampling of $n$ complete instances from the reference network; c) application of a MCAR missingness process, which turns each value of the instances into missing with probability $mp$;[2] d) execution of 30 runs of EM from different initializations, using the MAP-based score; e) choice of the estimate using MAP, entropy and BMA. To evaluate the quality of the estimates, we measure the KL-divergence between the joint distribution represented by the reference network and the estimated networks (denoted as *joint metric*). To measure the quality of the inferences produced by the models obtained with different criteria, it is necessary to select queries of interest. Having seen in experiments that both BMA and entropy yield consistently better estimates than MAP, we have chosen a query that could mitigate the differences among methods, in order to conservatively assess the advantage (in terms of inferences) of both BMA and entropy over MAP. In particular, we query the marginal joint distribution of all leaf nodes, without any evidence set in the network (*leaf metric*). This requires marginalizing out all non-leaf variables, so it involves all variables in the computation. Because of that, local "mistakes" in estimates can compensate each other, making harder to assess differences among methods. This is a desired characteristic if one wants to understand how strong is the difference among the ideas. Since KL-divergences are *not* normally distributed, we analyze the results through the non-parametric Friedman test with significance level of 1% (which is reasonably strong). To prevent issues from multiple comparisons, we performed the post-hoc of the test via Tukeys Honestly Significant Difference. Hence, the analysis yields a rank of methods for each setting and each metric. As for entropy, we choose the maximum entropy estimate among those whose score was at least as high as 95% of the highest score.

---

[1] Actually, BMA requires to average using as weights the posterior probability of each model, obtained as a product of its prior probability and its *marginal* likelihood, which implies in averaging over the parameters of the models.

## 3.1 ASIA Network

This set of experiments uses the structure of the Asia network [9]. In all settings (shown in Table 1) and the two metrics (joint and leaf), the Friedman test returned the rank: 1) BMA; 2) entropy; 3) MAP. To better understand the quantitative difference among them, we report in Table 1 the *relative medians* of KL divergence, namely the medians of BMA and entropy in a certain task, divided by the median obtained by MAP in the same task. This allows us to see the quantitative improvement of them over MAP.

The improvement of the median over MAP ranges, depending on the task, from 2% to 29% for BMA and from 1% to 14% for entropy. Moreover, the improvemenet is consistent, occurring in all settings. Interestingly, the difference in performance increase when the learning task is more challenging. For instance, the advantage of BMA over entropy, and of both BMA and entropy over MAP, increases with the percentage of missing data. Conversely, the greater the sample size the easier the learning task, thus the performance of the methods are more similar for $n=200$ than for $n=100$, even though the differences remain statistically significant in all the cases. As a result of the conservative design of the query, the differences among methods are generally less apparent in the leaf metric than in the joint metric. Overall, these experiments indicate BMA as the best option, while both BMA and entropy provide significantly better performance than MAP either in the parameter estimates and in the inferences. An insight of the reason for entropy to outperform MAP is given by Figure 1, which clearly shows that a higher MAP score does not necessarily imply in a better estimate; instead, when dealing with estimates of high MAP score, entropy is more discriminative than the MAP score and has also a stronger correlation with the KL divergence.

---

[2]MCAR (or *missing completely at random*) indicates that the probability of each value being missing does not depend on the value itself, neither on the value of other variables.

|  | n=100 | | | | n=200 | | | |
|  | mp=30% | | mp=60% | | mp=30% | | mp=60% | |
|  | BMA | entr. | BMA | entr. | BMA | entr. | BMA | entr. |
| --- | --- | --- | --- | --- | --- | --- | --- | --- |
| *joint* | 0.90 | 0.96 | 0.79 | 0.90 | 0.92 | 0.96 | 0.81 | 0.91 |
| *leaf* | 0.93 | 0.92 | 0.87 | 0.86 | 0.98 | 0.99 | 0.92 | 0.89 |

Table 1: Relative medians of KL divergence with the ASIA network, i.e., medians of BMA and entropy divided by the median of MAP. Smaller numbers indicate better performance; in particular, values smaller than 1 indicate a smaller median than MAP. Each cell corresponds to 300 experiments.

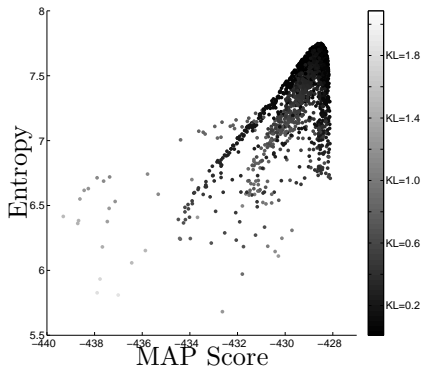

Figure 1: Relation between KL divergence, entropy and score; darker points represent lower KL divergence between true and estimated joint distributions. The figure refers to one thousand EM runs performed on an incomplete training set of 200 samples.

## 3.2 ALARM Network

The ALARM network has 37 nodes and 8 leaves [1]. Again, we consider $mp$ of 30% and 60%, and sample size $n$ of 100 and 200. As for the *joint* metric, in all settings the rank was: 1) BMA; 2) entropy; 3) MAP. As for the *leaf* metric, in all settings but one the rank was: 1) BMA & entropy; 2) MAP. The relative medians of KL divergence, reported in Table 2, show large differences on the joint metric than on the leaf one; only a slight difference exists between BMA and entropy in the latter case, while the advantage of both ideas over MAP is clearer. On the joint metric, BMA is by far the best-performing approach. Also in this case, the difference among ideas is emphasized when $mp$ increases or the sample size decreases. As before, BMA achieves the best results: it provides the better parameter estimates and it is at least as good as entropy for inferences. BMA and entropy consistently outperform MAP in the quality of parameter estimates and inferences.

|  | $n=100$ | | | | $n=200$ | | | |
|---|---|---|---|---|---|---|---|---|
| Metric | $mp=30\%$ | | $mp=60\%$ | | $mp=30\%$ | | $mp=60\%$ | |
|  | BMA | entr. | BMA | entr. | BMA | entr. | BMA | entr. |
| joint | 0.85 | 0.93 | 0.79 | 0.88 | 0.89 | 0.93 | 0.82 | 0.89 |
| leaf | 0.96 | 0.95 | 0.94 | 0.94 | 0.98 | 0.97 | 0.97 | 0.96 |

Table 2: Relative medians of KL divergence for experiments with the ALARM network. Each cell regards 300 experiments.

### 3.3 Randomly generated networks

In the case of randomly generated structures, the experimental procedure described in Section 3 also includes the generation of the random structure, which is accomplished before drawing the parameters. Given two variables $X_i$ and $X_j$, an arc from $X_i$ to $X_j$ is randomly included with probability 1/3 if $i < j$ (no arc is included if $j \geq i$, which ensures that the graph is acyclic and has no loops). Furthermore, the maximum number of parents is set to 4 and the number of categories per variable ranges from 2 to 4 (randomly chosen too). After the graph is generated, the experiments follow as before (see Table 3). On the *joint* metric, we always obtain the rank 1) BMA, 2) entropy, 3) MAP. On the *leaf* metric, we obtain that same rank in two settings and the rank 1) BMA & entropy; 2) MAP in the other two settings. Thus, there is a consistent superiority of BMA over entropy in estimating parameters, although this does not always implies in a superiority on the leaf queries. Both BMA and entropy are superior to MAP in all the cases.

In summary, these experiments indicate BMA as the best choice: it consistently yields the best parameter estimates, and on the leaf queries it is either the best idea or it ties with entropy. Overall, BMA and entropy are consistently better than MAP, both on the joint and on the leaf metric.

|       | n=100 |       |       |       | n=200 |       |       |       |
|       | mp=30% | | mp=60% | | mp=30% | | mp=60% | |
|       | BMA | entr. | BMA | entr. | BMA | entr. | BMA | entr. |
|-------|-----|-------|-----|-------|-----|-------|-----|-------|
| *joint* | 0.78 | 0.94 | 0.75 | 0.89 | 0.82 | 0.92 | 0.80 | 0.89 |
| *leaf*  | 0.89 | 0.92 | 0.88 | 0.88 | 0.96 | 0.97 | 0.95 | 0.92 |

Table 3: Relative medians for experiments with randomly generated networks with 20 nodes.

## 4 Experiments using Continuous Optimization

The maximum entropy criterion, applied to the selection of the parameter estimates, maximizes entropy while guaranteeing the score to exceed a certain threshold. So far, we have applied this idea on top of the multi-start EM, which selects only among estimates returned by the different EM runs. This approach identifies parameter estimates which are better than those from MAP, but usually worse than BMA's estimates. An alternative way to implement the idea of maximum entropy is to solve directly the non-linear optimization problem. This requires to maximize Eq. (2) subject to the constraint that the score is only marginally smaller than the best score, allowing a more fine-grained way to select the estimate than looking at the solutions identified by the different EM runs. This approach is referred in the following as *C_entropy*. The reason to analyze such situation is that EM runs tend to return local optima of the score function. However, the maximum entropy estimate might well be a non-optimum estimate in terms of score. Because of that, even if we increase the number of EM runs and use many more initializations (situation which we have tested), the entropy idea is still confined to saddle points of the score function, which is only an approximation of a true maximum entropy idea. Therefore, we perform experiments with two specific network structures with the aim of understanding whether the entropy idea improves by using a continuous optimization method instead of EM. For ease of expose, we focus only on the joint metric, as this is metric in which entropy is consistently inferior to BMA.

### 4.1 Experiments with BN1

Figure 2 shows the structure of the first set of experiments. Variables $A$ (binary) and $B$ (ternary) have uniform distributions and are always observed; variables $U$, $E$ and $T$ are binary (assuming states true and false); the value of $T$ is defined by the logical relation $T = E \wedge U$. Variable $T$ is always observed, while $U$ and $E$ are affected by the missingness process. Both $U$ and $E$ are observed if and only if $T$ is true. Therefore, $E$ and $U$ are either both observed and positive, or non-observed. The missingness process is MAR[3] because given $T$ (always observed) the probability of $U$ and $E$ to be missing does not depend on their values. Under the chosen conditions, $E$ and $U$ were missing in about 85% of the sampled instances. We assume the conditional probabilities of $T$ to be known, thus focusing on the difficulty of learning the probabilities of nodes $U$ and $E$.

---

[3]More precisely, this missingness process is MAR (as required by EM) but not MCAR (missing completely at random); for a discussion of the different kinds of missingness, see for instance [7, Sec. 19.1.2].

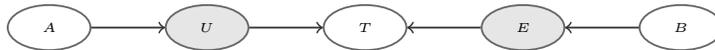

Figure 2: Network BN1; nodes affected by the missingness process have a grey background.

For this network, we also developed a solver which identifies the *global maximum* of the MAP score; technically, the solver maps the learning problem into a polynomial programming one; more details are given in the supplementary material. We use the global solver in two different ways: to compute *C_entropy*, thus ensuring the estimate to have a MAP score close to the global maximum (we allowed a tolerance of 1%); to compute the global MAP estimate on its own, referred to as *global MAP*. It is interesting to compare global MAP with MAP in order to assess which is the impact of the local solver (that is, EM) on the quality of estimates. To the best of our knowledge, there is no previous analysis with global exact solvers for learning BNs from incomplete samples.

The same experimental procedure of Section 3 is used, considering sample sizes in $\{100, 200\}$ and performing 300 experiments for each setting. For both sample sizes, the Friedman test on the joint metric returns the following rank: *1)* BMA; *2) C_entropy*; *3)* entropy; *4)* MAP; *5)* global MAP. The *relative* medians were, respectively, for $n$=100: 0.27, 0.33, 0.37, 1, 1.5; for $n$=300: 0.14, 0.25, 0.35, 1, 1.2. These results can be commented from several viewpoints. First, *C_entropy* significantly improves over entropy, almost reaching the same quality of BMA. Second, estimates found by the global solver were *worse* than those of MAP; in fact, the penalized MAP function offers only partial protection against overfitting; it is indeed less prone than log-likelihood to overfitting, but still an estimate which maximizes the MAP score can be well affected by overfitting. The value of MAP score achieved by the global solver was only slightly higher (around 2.5%) than achieved by the local maximum identified by the multi-start EM. Third, BMA and the two entropy implementations outperform MAP, confirming the results of the previous experiments.

## 4.2 Experiments with BN3

Network BN3 has structure $A \rightarrow B \rightarrow C$. We considered two different configurations of number of states for each node: 5-3-5 (meaning $A,C$ with 5 categories and $B$ with 3) and 8-4-8 ($A,C$ with 8 categories and $B$ with 4). In both cases, we made $B$ randomly missing on 85% of the instances. Thus, despite the simple structure of the network, the learning task is interesting since there are many missing data and a moderately high number of states. The two configurations requires to estimate *from incomplete samples* respectively $2 \cdot 5 + 4 \cdot 3 = 22$ and $8 \cdot 3 + 7 \cdot 4 = 52$ parameters; to these numbers, one should add the marginals of $A$, which are however learned from complete samples and whose estimate is thus identical for all methods. We fixed $n$ to 300 for the 5-3-5 configuration and to 500 for the 8-4-8. In this case, the performance of *C_entropy* was extremely good. We obtained, for both settings, the following rank in the Friedman test: *1)* *C_entropy*; *2)* BMA; *3)* entropy; *4)* MAP. The global MAP was not run, because the number of free unknowns in the continuous optimization problem was too high to achieve the global solution in reasonable time. The relative medians for the joint metric are (numbers are given in the order: *C_entropy*, BMA, entropy and MAP): for the 5-3-5 configuration: 0.71, 0.78, 0.90, 1; for the 8-4-8 configuration: 0.68, 0.78, 0.92, 1. These experiments confirm that, in order to get the most out of the entropy criterion, it is much more effective to have a dedicated solver than applying it on top of the multi-start EM. Here, *C_entropy* is even better than BMA, so its implementation for general settings and further analyses are intended in the near future.

## 5 Conclusions

This paper suggests that maximizing (penalized) likelihood or MAP scores is not the best choice to learn the parameters of a Bayesian network. In particular, a high score is necessary but not sufficient in order to have a good estimate. To improve estimation, we propose: (i) a BMA approach that averages over the estimates learned in different runs of EM, and (ii) a maximum entropy criterion applied over the estimates with high scores. The entropy idea can be implemented on top of EM or using a dedicated non-linear solver, which allows a more fine-grained choice of estimates. Both BMA and entropy can be promptly integrated into any EM implementation at virtually no cost in terms of implementation and running time; instead, the non-linear solver for entropy requires some additional implementation effort. Thorough experiments show that the presented ideas significantly improve the quality of estimates when compared to standard maximum penalized likelihood and MAP ideas. If EM is used as optimization engine, then BMA yields by far the best results, followed by entropy and MAP. If the dedicated non-linear solver is used, entropy performs as good as BMA, or even better. Moreover, for a specific network, we developed a global solver for the MAP estimation. We showed that its score is only slightly higher than the maximum identified by EM runs, and yet it yields worse estimates. This corroborates with the other results, indicating that the usual scores do suffer from overfitting and/or model uncertainty.